\documentclass[runningheads]{llncs}
\usepackage[T1]{fontenc}
\usepackage{booktabs}    
\usepackage{multirow}    
\usepackage[table]{xcolor} 
\usepackage{amsmath}
\usepackage{multirow}
\usepackage{bbding}

\usepackage{graphicx,verbatim}
\usepackage{graphicx}
\usepackage{orcidlink}
\begin{document}
\begin{sloppypar}
\newsavebox\CBox
\def\textBF#1{\sbox\CBox{#1}\resizebox{\wd\CBox}{\ht\CBox}{\textbf{#1}}}
\title{Symmetry Interactive Transformer with CNN Framework for 
Diagnosis of Alzheimer's Disease Using Structural MRI}
%
\titlerunning{SIT with CNN Framework} 
%
\author{Zheng Yang\inst{1}\orcidID{0009-0002-3710-7402} \and
Yanteng Zhang\inst{2}\orcidID{0000-0003-4796-2904} \and
Xupeng Kou\inst{3} \and
Yang Liu\inst{4} \and
Chao Ren\inst{1(}\Envelope\inst{)}}
\authorrunning{Z.Yang et al.}
%
\institute{College of Electronics and Information Engineering, Sichuan University, China \and
Tri-Institutional Center for Translational Research in Neuroimaging and Data Science (GSU, Georgia Tech, Emory), Atlanta, USA \and
College of Information and Electrical Engineering, China Agricultural University, China \and
School of Biomedical Engineering and Imaging Sciences, King's College London, UK\\
\email{chaoren@scu.edu.cn}\\
}
\maketitle              

\begin{abstract}
Structural magnetic resonance imaging (sMRI) combined with deep learning has achieved remarkable progress in
the prediction and diagnosis of Alzheimer’s disease (AD).
Existing studies have used CNN and transformer to build a well-performing network, but most of them are based on pretraining or ignoring the asymmetrical character caused by brain disorders. We propose an end-to-end network for the detection of disease-based asymmetric induced by left and right brain atrophy which consist of 3D CNN Encoder and Symmetry Interactive Transformer (SIT). Following the inter-equal grid block fetch operation, the corresponding left and right hemisphere features are aligned and subsequently fed into the SIT for diagnostic analysis. SIT can help the model focus more on the regions of asymmetry caused by structural changes, thus improving diagnostic performance. We evaluated our method based on the ADNI dataset, and the results show that the method achieves better diagnostic accuracy (92.5\%) compared to several CNN methods and CNNs combined with a general transformer. The visualization results show that our network pays more attention in regions of brain atrophy, especially for the asymmetric pathological characteristics induced by AD, demonstrating the interpretability and effectiveness of the method.

\end{abstract}
\section{Introduction}
AD (Alzheimer’s disease) is a progressive neurodegenerative disease that severely affects cognitive function and daily living ability of patients. The early symptoms of AD, known as mild cognitive impairment (MCI), have a minimal impact on daily behavioral abilities, but more than 30\% of patients with MCI progress to AD in 5 years~\cite{ref_article9}. Early diagnosis and intervention are critical for slowing disease progression and improving patient quality of life~\cite{ref_article3}. In recent years, structural Magnetic Resonance Imaging (sMRI) data combined with deep learning techniques have made significant progress in the prediction and diagnosis of AD and its prodromal stage [ie MCI]\cite{ref_lncs15,ref_lncs16,ref_lncs17}. sMRI provides high-resolution images of brain structures, from which deep learning techniques can extract complex features to recognize and classify diseases. 

Symmetry between the left and right brain hemispheres serves as a crucial biomarker in numerous neurological and psychiatric disorders, playing a pivotal role in both diagnostic evaluation and disease progression monitoring, while often providing early indicators of pathological changes\cite{ref_article12,ref_article13}. Measurements of pathological structural and functional asymmetries not only aid in the early recognition of disease but can also be used to manage disease progression and assess the effectiveness of treatment. Such asymmetric alterations are particularly pronounced in diseases such as AD~\cite{ref_article1} and schizophrenia, where hemispheric symmetry is considered by radiologists as one of the most indicative biomarkers~\cite{ref_lncs4}. However, most of the existing deep learning methods focus on the extraction and analysis of unilateral brain hemispheric features, either directly on the whole brain in general, ignoring information about interactions and asymmetries between the left and right hemispheres, or only for preprocessing and data enhancement rather than encoding within a network structure, which limits the ability of generalization to be extended to the general process of neuroimaging analysis.

In order to overcome this limitation and take full advantage of the symmetry of the left and right brain, our study proposes a novel approach that can utilize the symmetry between right and left brain hemispheres. We adapt the transformer model based on Zhang et al.~\cite{ref_lncs2} work to capture the basic similarity. To obtain the whole symmetry and asymmetry features, we adopt the HemiSim Matrix to amplify the directional similarity~\cite{ref_lncs3}. Subsequently our proposed Symmetry Interactive Transformer(SIT) can focus more attention on Disease-induced asymmetrical regions. This study not only enriches the existing diagnostic techniques for AD, but also provides new perspectives for understanding the structural asymmetry of the brain in neurodegenerative diseases. It is hoped that this study will yield breakthroughs in the early diagnosis and treatment of AD. The main contributions of this study are as follows:

\begin{enumerate}
    \item We design a framework with CNNs extracting hemisphere-specific features and a Symmetry Interactive Transformer (SIT) for cross-hemispheric fusion. This design synergizes local-global structural features of brain, significantly improving neuroimaging diagnostic accuracy.

    \item By introducing the HemiFuse Attention mechanism of the transformer model, we achieve efficient interaction between left and right brain features, further enhancing the expressive power of these features.
    
    \item We construct the HemiSim Matrix by utilizing the symmetry difference between the right and left hemispheres of diseased and normal individuals.

\end{enumerate}
\section{Methodology}
\begin{figure}[htbp]
    \centering
    \includegraphics[width=1\textwidth]{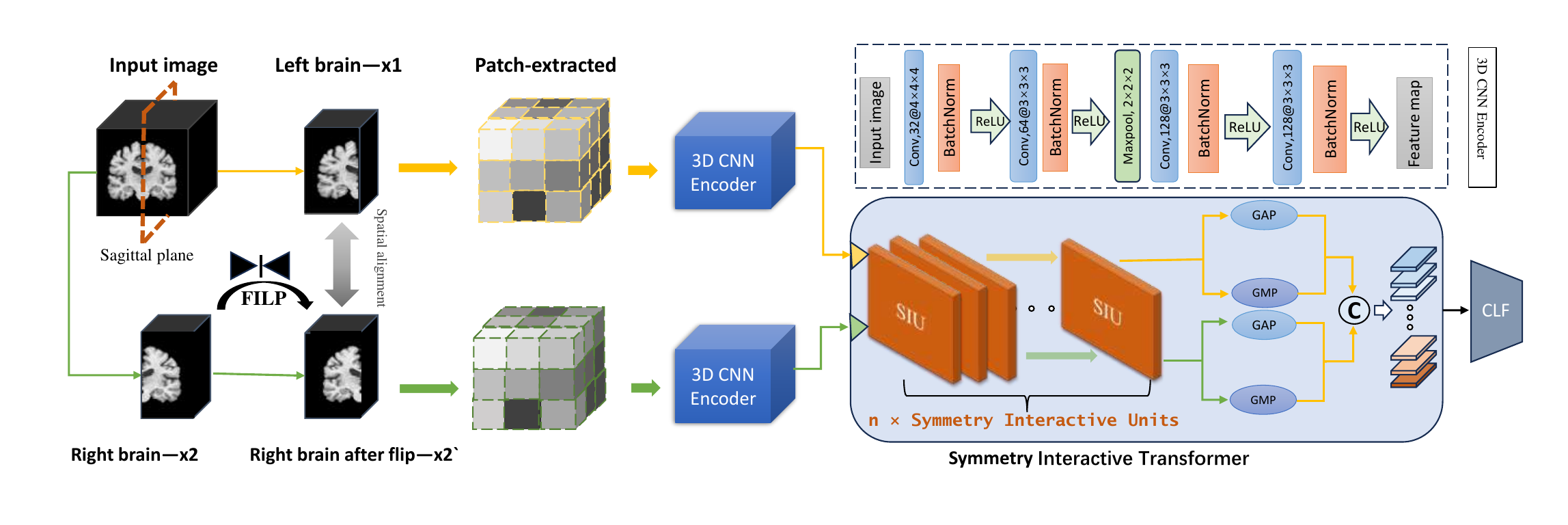} 
    \caption{ Overview of our proposed network. SIU denotes the Symmetry Interactive Unit composing the Symmetry Interactive Transformer. CLF denotes the classifier for final prediction. GAP and GMP denote global average pooling and global max pooling, respectively.} 
    \label{fig:111} 
\end{figure}

The proposed model is composed of four components, as illustrated in Fig.~\ref{fig:111}. These components include an input block fetcher, patch-level 3D CNN encoders, a transformer encoder, and a classifier. The sMRI scan, obtained from the subject, is initially segmented into left and right brain components according to the sagittal plane, with the right brain being flipped along the sagittal plane to align with the left brain~\cite{ref_lncs4}. Subsequently, the blocks are extracted according to the isometric grid, ensuring that each block entering the CNN network for feature extraction corresponds to the symmetric position of the left and right brain. The extracted left and right brain chunks are then fed into the Symmetry Interactive Transformer encoder (composed of n Symmetry Interactive Units) to interact with the features and weight the left and right brain similarities to learn the asymmetry of the left and right brain due to the structural changes. Unlike the traditional Vision Transformer(ViT) model, our proposed model combines the CNN with the transformer to address the poor performance of ViT on small datasets and the shortcomings of the CNN's inability to capture long-range block relationships in images.
\subsection{3D CNN Encoder}
In order to efficiently process spatial information of the patch in the left and right hemispheres, the 3D CNN encoder~\cite{ref_article10} was used to extract every patch features. The encoder consists of four convolutional modules, each including a 3D convolutional layer, a batch normalization (BN) layer, and a ReLU activation function. The first convolutional layer utilizes a 4×4×4 kernel, enabling the capture of a comprehensive sensory field and global features. The subsequent three convolutional layers employ a 3×3×3 kernel, a reduction in complexity and parameter size, while preserving local image details. The input has the shape [B*N,1,H,W,D], which is processed by the 3D CNN encoder to obtain a feature map with the shape [B*N,C,H,W,D], where C=128 is the number of feature channels.
\subsection{Symmetry Interactive Transformer}
\subsubsection{Transformer Block} 
\begin{figure}[htbp]
    \centering
    \includegraphics[width=1\textwidth]{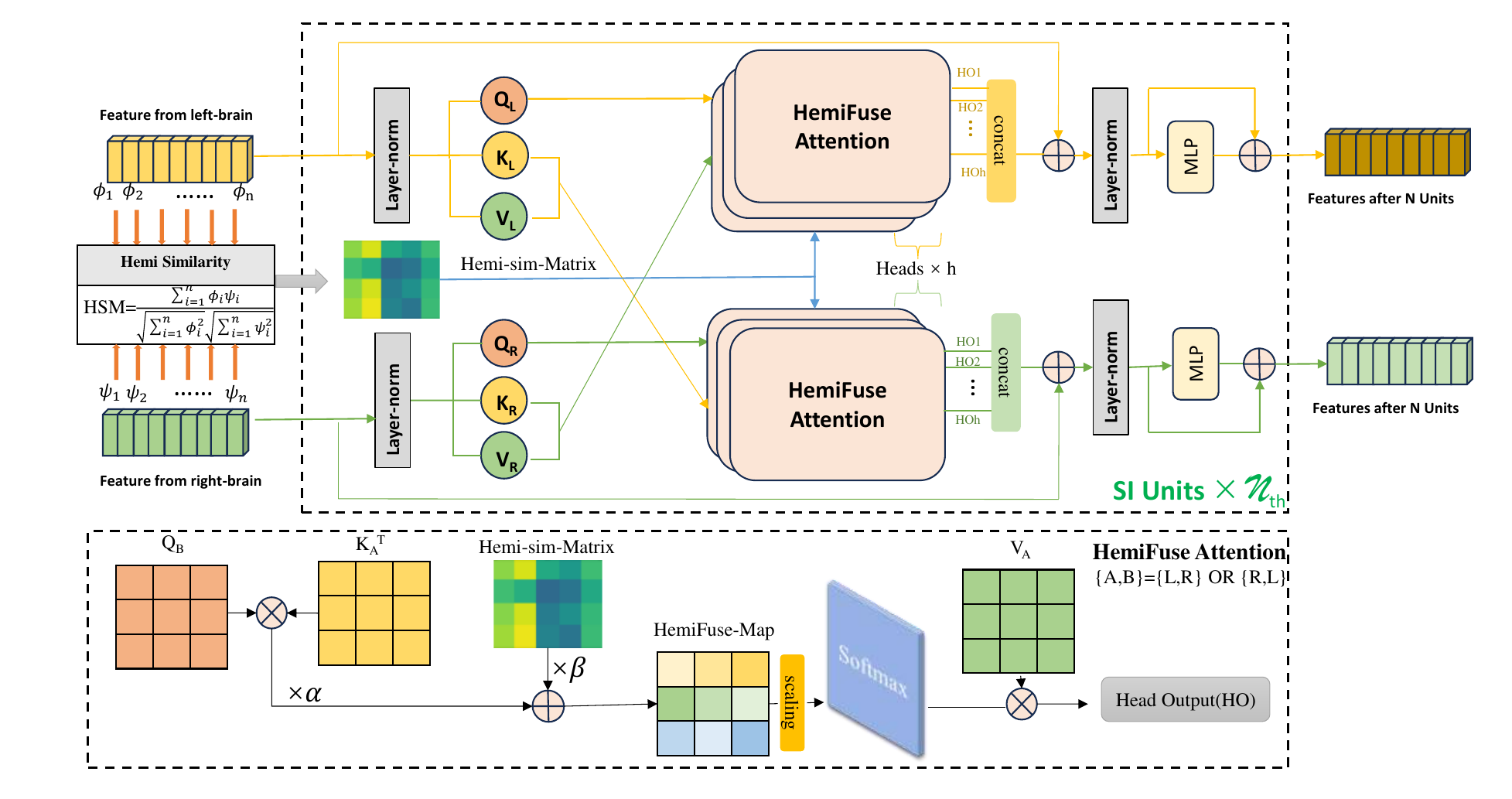} 
    \caption{ Detailed structure of our proposed Symmetry Interactive Unit. HSM denotes HemiSim Matrix calculated by eq\eqref{eq:hsm}. \(\alpha\) and \(\beta\) denote the learnable parameters.} 
    \label{fig:333} 
\end{figure}

This work introduces a HemiFuse Attention inspired by the cross-attention mechanism~\cite{ref_lncs7} in order to obtain the similarity computation of the left and right brains as shown in Fig.~\ref{fig:333}. The left and right brain feature maps generated by the CNN encoder are transformed into feature vectors. The feature vectors of the left and right brains are subsequently mapped into Q, K, and V. In HemiFuse Attention, Q from one side and K and V from the other side are subjected to the eq\eqref{eq:attn}, while in the computation of the attention scores. \\

The model incorporates a weighting module based on the HemiSim Matrix of the left and right brains. A multi-head strategy is employed for parallel computation of cross-attention, allowing the model to focus on distinct information at varying levels. The operational dynamics of the Transformer encoder can be articulated through the ensuing formula:
\begin{equation}
\begin{split}
F &= Attention + f,\\
G &= MLP(LN(F))+ F,
\end{split}
\end{equation}
where f denotes the input feature vector, LN is used as layer normalization,  MLP represents the fully connected layer and G denotes the output from transformer.
\subsubsection{HemiSim Matrix (HSM)} We added the HemiSim Matrix between each corresponding block feature of the left and right brain, which is calculated as the cosine similarity between the features extracted by the CNN in the feature channel dimension with the following formula:
\begin{equation}
HSM =\frac{\sum_{i=1}^{n} \phi_i \psi_i}{\sqrt{\sum_{i=1}^{n} \phi_i^2} \cdot \sqrt{\sum_{i=1}^{n} \psi_i^2}},\
\label{eq:hsm}
\end{equation}
where \(\phi_i\) and \(\psi_i\) denote the $i$-th input feature from left and right brain. The input feature shape is defined as [B,N,C], where B denotes the batch size, N denotes the number of left / right brain chunks (equivalent to the length of the feature sequence), and C represents the number of feature channels. This configuration results in a matrix of shape [B,N]. This matrix has the capacity to enhance the focus on directionality in attention and improve the detection of symmetry of the left/right brain. Therefore, the new HemiFusion Attention calculation formula is as follows:
\begin{equation}
HemiFusion Attention = softmax(\frac{\alpha(Q*K^T)+\beta HSM}{\sqrt{d_h}})*V,
\label{eq:attn}
\end{equation}
where Q denotes the query from one side, K and V represent the key and value from the other side.  Two learnable scalar weights \(\alpha\) and \(\beta\) are introduced to adaptively balance the contributions of the standard dot-product attention and HemiSim Matrix. \(d_{h}\) denotes the dimension of each attention head.
\subsection{Implement Details}
\subsubsection{Classifier}  The function of the classifier after SIT can be expressed by the following equation:
\begin{equation}
Pre = softmax[LN(ReLU(LN(g))],
\end{equation}
where LN denotes a linear transformation that maps features to a hidden layer and then to the final classification output, and a softmax is used to convert the output of the last hidden layer to a predicted probability distribution. g denotes the tokens from SIT.
\subsubsection{Loss}We use three cross-entropy loss functions jointly in a set ratio as the training loss function.
\begin{equation}
LOSS = \lambda*loss_L + \gamma*loss_R + loss_\text{all},
\label{eq:loss}
\end{equation}
where \(loss_L\) and \(loss_R\) denote the loss function of the predicted values of each side of the left and right brain obtained through the output of the classifier after convolution, respectively. \(\lambda\) and \(\gamma\) are weight parameters introduced to balance the contribution between \(loss_L\) and \(loss_R\) in LOSS.
\(loss_\text{all}\) denotes the final prediction value output by the model at the end. By simultaneously optimising the left and right brain and the final prediction performance, the joint cross-entropy loss function can enable the model to learn more comprehensive features and improve the robustness of the model.

\section{Experiments And Results}
\subsection{Dataset}
ADNI (Alzheimer's Disease Neuroimaging Initiative)~\cite{ref_article6} is currently the most widely publicised AD research dataset. It was used for this experiment, and the data for the AD study is comprised of 801 medical images, 213 of which are classified as AD and 234 as CN (Cognitively Normal) and 219 sMCI and 135 pMCI subjects.  We divide the imaging dataset proportionally into 70\% of the training set, 15\% of the validation set and 15\% of the test set.

At the same time, we preprocessed brain images on a clinical platform~\cite{ref_article7} following standard procedures. Specifically, the sMRI is first subjected to pre-combined post-combined correction, followed by affine alignment with a standardized template in MNI152 space. Finally, the pre-processed sMRI images are subjected to skull stripping~\cite{ref_article8}. The resolution of the preprocessed sMRI scans is 121×145×121.

\subsection{Experimental Settings}
All the experiments are conducted on a NVIDIA RTX 4060Ti-16GB GPU, and during the model training process, we employ the Adam algorithm to optimize the parameters. We set a batch size of 4 and the epoch to 70. During the training process, the learning rate of the first 30 training epochs is set to 1e-4, the 30th to 50th training epochs are set to 3e-5, and the learning rate is kept at 1e-5 for the last following epochs.
\subsection{Ablation Study Result}
We performed several sets of ablation experiments as following Table~\ref{tab1} and each using the same CNN backbone, block fetching operation, and SI Transformer as used in our proposed model.
\begin{table}[htbp]
\centering
\caption{Ablation experimental results on ADNI test set (\%)}\label{tab1}
\small
\setlength{\tabcolsep}{4pt}
\renewcommand{\arraystretch}{1.1}
\begin{tabular}{@{} l *{4}{r} *{4}{r} @{}}
\toprule
\multirow{2}{*}{Method} & \multicolumn{4}{c}{AD vs CN} & \multicolumn{4}{c}{pMCI vs sMCI} \\
\cmidrule(lr){2-5} \cmidrule(l){6-9}
& ACC & SEN & SPE & AUC & ACC & SEN & SPE & AUC \\ 
\midrule
1-wCNN & 80.60  & 93.75  & 68.57  & 81.16  & 69.81 & 85.00 & 60.61 & 72.80  \\
2-LRCNN & 86.57  & 84.38  & 88.57  & 86.47  & 73.58 & 65.00 & 78.79 & 71.90  \\
3-LRCNN-T & 86.57  & 84.38  & 88.57  & 86.47  & 75.47 & 75.00 & 75.76 & 75.38  \\
4-LRCNN-IT & 88.06  & 84.38  & 91.43  & 87.90  & 75.47 & 75.00 & 75.76 & 75.38  \\
5-LRCNN-SIT & 85.07  & 87.50  & 82.86  & 85.18  & 71.69 & 40.00 & 90.91 & 65.45  \\
\rowcolor{gray!10}
6-LRCNN-SIT(FLIP) & \textBF{92.54} & 87.50 & \textBF{97.14} & \textBF{92.32} & \textBF{77.36} & 75.00 & 78.79 & \textBF{76.89}  \\
\bottomrule
\end{tabular}
\end{table}

The first group wCNN was classified by just convolving the whole brain with a CNN network. the second group LRCNN was classified by convolving the left and right brains with separate CNN networks, the third group LRCNN-T was based on the second group with the addition of the transformer but without the left and right brain interactions, i.e., the Q, K, and V of the dotted-accumulation of the attention came from the same side of the brain; the fourth group LRCNN-IT(interactive transformer)used the left and right brain interactions of the transformer; the fifth group LRCNN-SIT(symmetry interactive transformer) added the sim matrix on addition to the interaction, and the sixth group mirrored the right brain to flip it so that its orientation was perfectly aligned with the left brain, i.e., the final complete network proposed in this paper.

From six sets of increasingly comprehensive ablation experiments, it can be observed that each improvement we proposed has enhanced the model’s performance, except for the fourth and fifth groups. In the fifth group, a sim matrix is incorporated based on the fourth group with the expectation of obtaining the symmetry features of the left and right brain hemispheres. However, the absence of  FLIP operation caused the model to acquire some incorrect features, resulting in a decline in performance. When  FLIP operation was added (as in the sixth group), there was a significant improvement in performance.
\begin{figure}[htbp]
    \centering
    \includegraphics[width=1.0\textwidth]{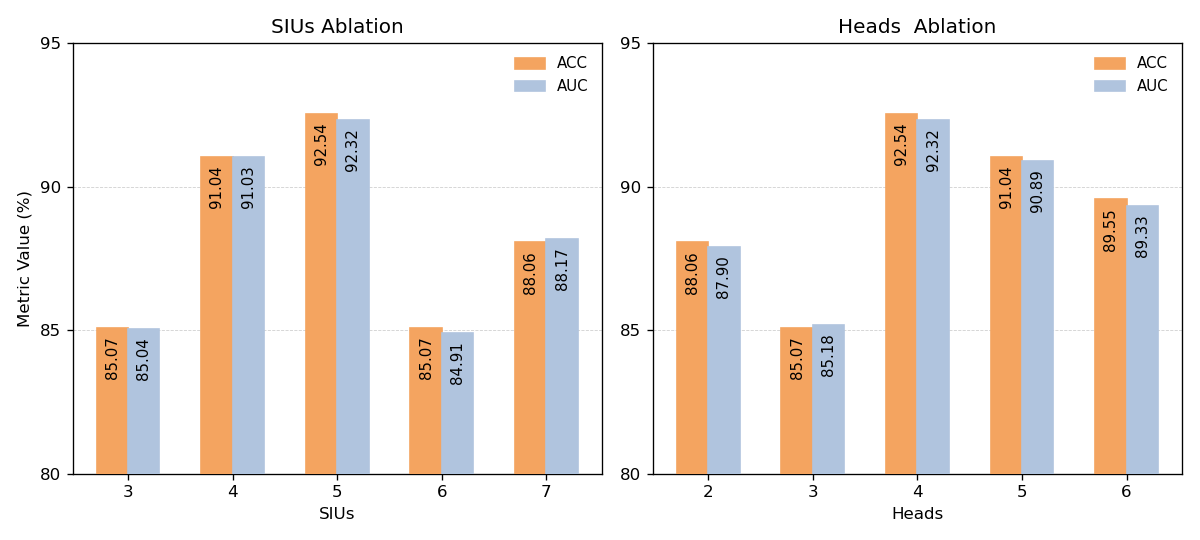} 
    \caption{ Impact of numbers of SI Units and Heads, on the proposed method in AD prediction. } 
    \label{fig:num} 
\end{figure}
\subsection{Parameter Analysis}

\subsubsection{SIT Parameter} To get the best diagnosis performance on SIT, we perform many sets of experiments to find the most suitable parameter configuration. The parameters of the SI Transformer have a huge impact on the whole network, in which we perform a series of ablations of the number of SI Units, the number of heads. From Fig.~\ref{fig:num} we can see that SI Transformer can yield good performance with SIUs = 5, Heads = 4.
\subsubsection{Patch-size}Prior to CNN-based feature extraction, sMRI images are partitioned into fixed-size 3D blocks using an isometric grid layout. This patching strategy ensures that each block from one hemisphere has a corresponding symmetric counterpart in the other hemisphere, enabling symmetric feature comparison. To ensure that the entire brain volume is evenly covered by patches, zero-padding is applied to the original image when its dimensions are not divisible by the selected patch size. The size of each patch plays a critical role in determining the granularity of local feature extraction and the overall model complexity. So we further compare the effect of patch size on model performance and complexity, as shown in Table~\ref{tab1}. 
\begin{table}[htbp]
\centering
\caption{Effect of Patch Size on AD vs CN and pMCI vs sMCI Classification on the ADNI Test Set (\%).}
\label{tab1}
\small
\setlength{\tabcolsep}{6pt}
\renewcommand{\arraystretch}{1.15}
\begin{tabular}{l c  cc cc}  
\toprule
\multirow{2}{*}{Patch Size} & \multirow{2}{*}{Params (M)}  & \multicolumn{2}{c}{AD vs CN} & \multicolumn{2}{c}{pMCI vs sMCI} \\
\cmidrule(lr){3-4} \cmidrule(lr){5-6}
& & ACC & AUC & ACC & AUC \\
\midrule
15 & 5.80  & 88.06  & 88.30  & 75.47  & 75.38   \\
20 & 4.47  & 86.58  & 86.74  & 73.58  & 71.90   \\
\rowcolor{gray!10}
25 & 3.77  & \textBF{92.54}  & \textBF{92.45} & \textBF{77.36}  & \textBF{76.89}   \\
30 & \textBF{3.70}  & 86.56  & 86.60  & 75.47  & 75.38   \\
\bottomrule
\end{tabular}
\end{table}

It can be observed that the model achieves the best performance when the patch size is set to 25, reaching the highest accuracy of 92.54\% and AUC of 92.45\%, while also maintaining a relatively low parameter count (3.77M). Smaller patch sizes lead to increased model complexity without substantial performance gain, whereas larger patch sizes slightly reduce parameters but suffer from accuracy degradation. These results suggest that a moderate patch size provides a good trade-off between performance and model complexity.
\subsubsection{Loss Parameter}
\begin{figure}[htbp]
    \centering
    \includegraphics[width=1.0\textwidth]{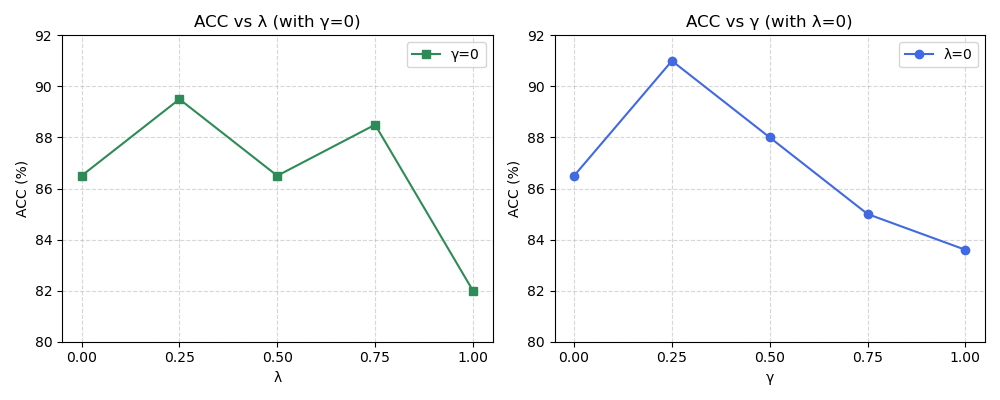} 
    \caption{ Impact of two parameters \(\lambda\) (with \(\gamma\) = 0) and \(\gamma\) (with \(\lambda\) = 0), on the proposed method in AD prediction. } 
    \label{fig:beta} 
\end{figure}

The parameters \(\lambda\) and \(\gamma\) in  Eq\eqref{eq:loss} play an important role in balancing the contributions of the left and right brain hemispheres with the whole brain. To study their influence on the proposed SIT model, we tune their values within the range of [0, 0.25, 0.50, 0.75, 1.00], and report the corresponding ACC values in AD prediction task. From Fig.~\ref{fig:beta}, we can see that SIT can yield good performance with \(\lambda\) = \(\gamma\) = 0.25. Additionally, with \(\lambda\) = 0 and \(\gamma\) = 0, the ACC values are not good. This suggests that \(loss_L\) and \(loss_R\) have the same positive complementary effects on improving the diagnosis.
\subsection{Comparison Study Result}
In order to verify the validity and feasibility of our improved model, we chose other recent excellent AD diagnostic works with ADNI datasets in comparable numbers.
\begin{table}[htbp]
\caption{Comparison experimental results for AD vs. CN on the ADNI test set(\%)}\label{tab2}
\centering
\small
\setlength{\tabcolsep}{4pt}
\renewcommand{\arraystretch}{1.1}
\begin{tabular}{@{} l l l c c c c @{}}
\toprule
Study & Method & Publication & ACC & SEN & SPE & AUC \\ 
\midrule
Zhang~\cite{ref_lncs2} & CNNVIT & ISBI 2023 & 0.881 & 0.844 & 0.914 & 0.879 \\
Liu~\cite{ref_lncs5} & MTFIL-Net & BIBM 2021 & 0.860 & 0.810 & 0.900 & 0.950 \\
Zhang~\cite{ref_zh} & ATTCNN & ISBI 2024 & 0.891 & 0.844 & 0.933 & 0.89 \\
Lian~\cite{ref_article4} & AGHyb-Net & TCYB 2022 & 0.919 & 0.887 & 0.945 & 0.965 \\
Zhu~\cite{ref_article5} & DA-MIDL & TMI 2021 & 0.924 & 0.910 & 0.938 & 0.965 \\
Jiang~\cite{ref_lncs6} & AAGN & MICCAI 2024 & 0.903 & --- & --- & 0.947 \\
\rowcolor{gray!10}
Ours & LRCNN-SIT & --- & \textBF{0.925} & 0.875 & \textBF{0.971} & 0.923 \\
\bottomrule
\end{tabular}
\end{table}
Zhang~\cite{ref_lncs2} uses a similar framework as CNN encoders combined with a cross-attention transformer to process MRI and PET modalities. Liu~\cite{ref_lncs5} has implemented AD detection and MMSE score via the CNN network. Zhang~\cite{ref_zh} improves diagnostic performance by exploiting whole-brain and gray matter images through attention-guided CNN. Lian~\cite{ref_article4} uses the same block size for CNN feature extraction for AD detection as well. Zhu~\cite{ref_article5} proposes CNN networks combining patch and attention. Jiang~\cite{ref_lncs6} Composes Gated Networks via CNN Decoders and Attention Mechanisms to Enhance AD Diagnostics. As shown in the Table~\ref{tab2}, our method can achieve very good performance compared with other excellent methods, which fully demonstrates the feasibility and effectiveness of our proposed method.

\subsection{Visualization Results}
\begin{figure}[htbp]
    \centering
    \includegraphics[width=1.0\textwidth]{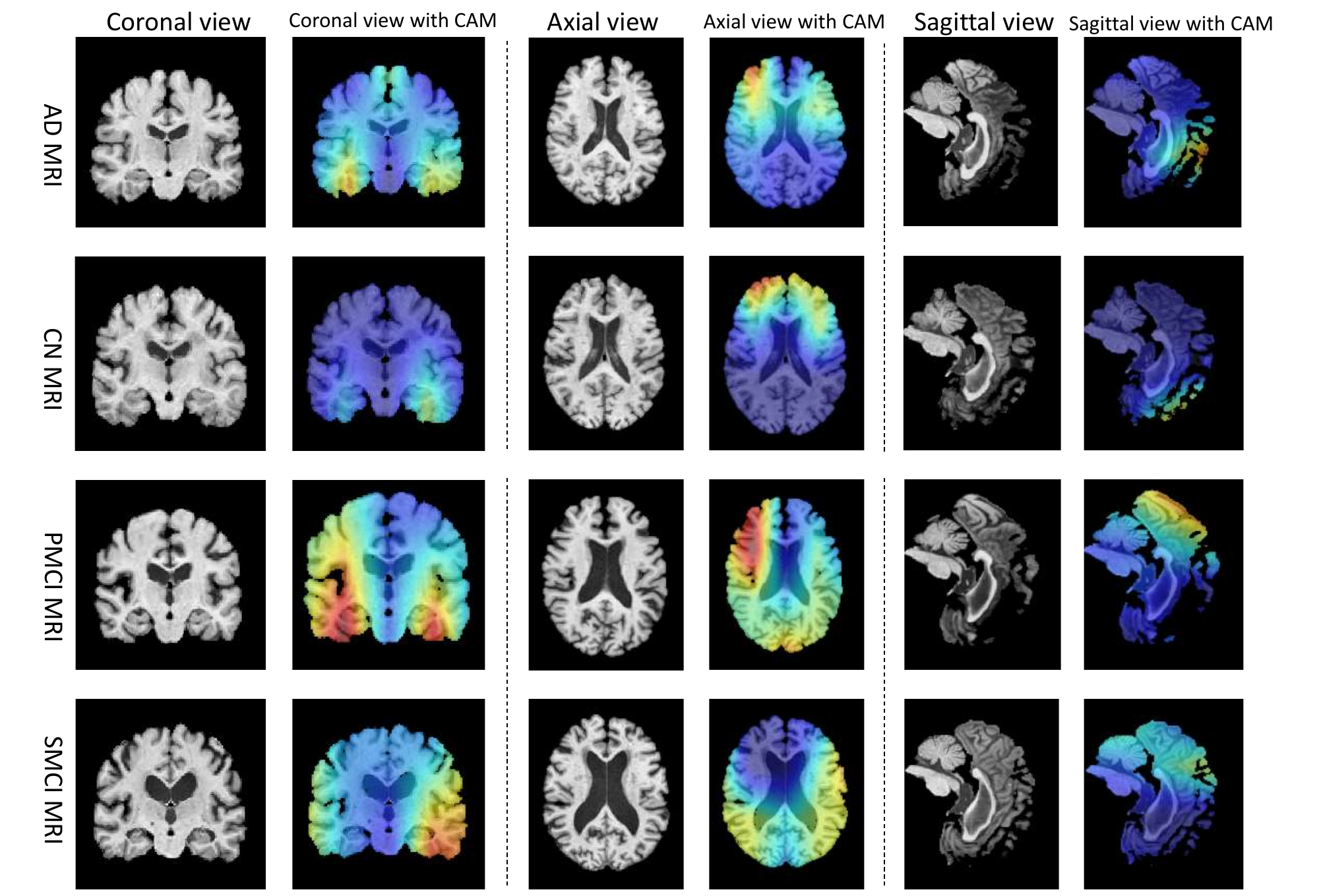} 
    \caption{ Visual interpretation of the heat maps on brain images. } 
    \label{fig:555} 
\end{figure}

We choose the last layer of weights of SIT to be visualized by the 3D Grad CAM technique, and from the results, our proposed model is able to identify the symmetric characteristics of the brain related to dementia disease very well. For example, the hippocampus regions~\cite{ref_article11} in the sMRI are distorted due to AD and MCI, creating a significant asymmetry with the other side. From Fig.~\ref{fig:555} we can see that SIT focused on this abnormality and assisted in the final prediction accordingly, resulting in an improved diagnostic result.
\section{Conclusion}
In this study, we proposed a novel framework for AD diagnosis by leveraging the distinct pathological changes of the left and right hemispheres through CNN combining the Symmetry Interactive Transformer. By independently extracting features from each hemisphere, our approach enables the observation of structural abnormalities, which in turn lead to varying degrees of asymmetry commonly associated with AD and MCI. The results of the ablation experiments confirm the effectiveness and feasibility of the proposed method. Furthermore, visualization of the learned features clearly reveals the network's ability to localize brain atrophic regions, especially in the case of brain dementia diseases with asymmetric pathology. These results underscore the potential of our model as a valuable tool for supporting clinical decision-making in the detection and analysis of AD.
\begin{credits}
\subsubsection{\ackname} This work was supported by the Young Faculty Technology Innovation Capacity Enhancement Program of Sichuan University under Grant 2024SCUQJTX025. Data used in preparation of this article were obtained from the Alzheimer's Disease Neuroimaging Initiative (ADNI) database. As such, the investigators within the ADNI contributed to the design and implementation of ADNI and/or provided data but did not participate in analysis or writing of this report. More details can be found at adni.loni.usc.edu.

\end{credits}

%
%
%
%
\bibliographystyle{splncs04_unsort}
\bibliography{references}
\end{sloppypar}
\end{document}